\documentclass[manuscript, screen]{jair}

\setcopyright{cc}
\acmDOI{10.1613/jair.1.xxxxx}

\JAIRAE{Insert JAIR AE Name}
\JAIRTrack{}
\acmVolume{0}
\acmArticle{0}
\acmMonth{3}
\acmYear{2026}

\RequirePackage[
  datamodel=acmdatamodel,
  style=acmauthoryear,
  backend=biber,
  giveninits=true,
  uniquename=init
]{biblatex}

\usepackage{booktabs}
\usepackage{multirow}
\usepackage{amsmath}

\begin{document}

\title{Hallucination as Output-Boundary Misclassification: A Composite Abstention Architecture for Language Models}

\author{Angelina (Davini) Hintsanen}
\authornote{Corresponding Author.}
\email{angelina@nexus-ailab.org}
\affiliation{%
  \institution{NEXUS Laboratory}
  \city{Helsinki}
  \country{Finland}
}

\renewcommand{\shortauthors}{Hintsanen}

\begin{abstract}
Large language models routinely produce unsupported claims---a failure termed hallucination. We propose a control-theoretic framing: hallucination is a misclassification error at the output boundary, where internally generated completions are emitted as if grounded in evidence. This framing motivates a composite intervention combining instruction-based refusal with a structural abstention gate. The gate computes a support deficit score $S_t$ from three black-box signals---self-consistency ($A_t$), paraphrase stability ($P_t$), and citation coverage ($C_t$)---and blocks output when $S_t$ exceeds a threshold. In a controlled evaluation across 50 items, five epistemic regimes, and three models (GPT-4o-mini, GPT-4o, GPT-3.5-turbo), neither mechanism alone was sufficient: instruction-only prompting reduced hallucination sharply, but exhibited over-cautious abstention on 10\% of answerable items for GPT-4o-mini and GPT-4o, and residual hallucination for GPT-3.5-turbo (6\% overall; driven primarily by conflicting-evidence items). The structural gate preserved 100\% answerable accuracy across models but missed confident confabulation on conflicting-evidence items (70\% hallucination for GPT-4o-mini and GPT-4o). The composite architecture achieved 96--98\% overall accuracy with 0--4\% hallucination, while also inheriting the instruction component's 10\% abstention on answerable items for GPT-4o-mini and GPT-4o. A supplementary 100-item no-context stress test derived from TruthfulQA confirmed that structural gating provides a capability-independent abstention floor: instruction-only abstention degraded to 62\% for GPT-3.5-turbo, whereas the gate and composite conditions enforced 98--100\% abstention across all models. These results are consistent across the three tested models, though architecture-level generality remains to be established. Overall, instruction-based refusal and structural gating exhibit complementary failure modes---instruction can over-abstain on answerable items, while the gate can miss confident confabulation under conflicting evidence---suggesting that effective hallucination control benefits from combining both mechanisms.
\end{abstract}

\received{9 March 2026}
\maketitle

\section{Introduction}
\label{sec:intro}

Large language models (LLMs) generate text by predicting the next token conditional on preceding context \citep{brown2020language, touvron2023llama}. This autoregressive process produces fluent output but routinely generates claims not supported by input evidence---termed hallucination \citep{ji2023survey, huang2023survey, maynez2020faithfulness}. Hallucination persists under retrieval augmentation \citep{shuster2021retrieval}, instruction tuning, and reinforcement learning from human feedback \citep{ouyang2022training}.

\subsection{The Gap-Filling Problem}
\label{sec:gap}

When a query requires information absent from the prompt, retrieved context, and the model's stored parametric knowledge, the system faces an epistemic gap. The model continues generating because training rewards fluent completion rather than epistemic caution \citep{lin2022truthfulqa, kadavath2022language}. We propose that this gap-filling behavior constitutes a misclassification: when prior-driven completion is emitted without being distinguished from evidence-backed generation, the system treats unsupported continuation as grounded output.

This framing is conceptually inspired by control-theoretic models of biological inference, where internally generated signals (e.g., interoceptive arousal during simulation) may be misclassified as confirmatory external evidence rather than treated as internally generated states requiring regulation, sustaining problematic positive-feedback loops. The analogous loop in LLMs is:
\begin{equation}
\text{query} \rightarrow \text{gap} \rightarrow \text{prior-only completion} \rightarrow \text{emitted as answer} \rightarrow \text{user accepts}
\end{equation}

The intervention target is classification at the output boundary.

\subsection{From Post-Hoc Detection to Pre-Output Control}
\label{sec:preoutput}

Dominant mitigation strategies operate after generation: checking output against sources \citep{manakul2023selfcheckgpt, min2023factscore}, training verifiers \citep{lightman2024verify}, or using self-consistency voting \citep{wang2023selfconsistency}. These share a structural limitation: hallucinated content has already been produced. Selective prediction \citep{elyaniv2010foundations, geifman2017selective} offers an alternative: a pre-output gate that blocks generation when epistemic support is insufficient.

Existing hallucination work often characterizes the problem at the level of observable output: false content, unsupported claims, or benchmark failure. The present paper adopts a narrower structural question: under what conditions does internally generated completion become treated as if it were evidentially sufficient response? Framed this way, hallucination is not only a content error but a boundary error in the classification of support. This reframing shifts attention away from post hoc characterization of what went wrong and toward pre-emission control of when generation should be withheld.

\subsection{Contributions}
\label{sec:contributions}

\begin{enumerate}
    \item A control-theoretic framing of hallucination as output-boundary misclassification.
    \item A black-box support-deficit score derived from three externally measurable signals.
    \item Empirical evidence from 50 items $\times$ 3 models $\times$ 4 conditions demonstrating that a composite architecture (instruction + gate) achieves 96--98\% accuracy with 0--4\% hallucination under the tested conditions.
    \item A supplementary 100-item no-context stress test derived from TruthfulQA confirming that structural gating provides a capability-independent abstention floor when instruction-following degrades.
    \item Identification of the specific failure mode of structural gating (confident confabulation) and the specific failure mode of instruction-only prompting (over-cautious abstention on answerable items for GPT-4o-mini/4o; residual hallucination for GPT-3.5).
\end{enumerate}

The experiments presented here are designed as a controlled proof-of-concept evaluation of the proposed architecture rather than a large-scale benchmark comparison. The goal is to test whether the composite abstention mechanism addresses complementary failure modes of instruction-based refusal and structural gating under clearly defined epistemic regimes.

An earlier version of this architecture was accepted to the ICLR 2026 Workshop on LLM Reasoning \citep{davini2026hallucination}.

\section{Theoretical Framework: Output-Boundary Misclassification}
\label{sec:theory}

\subsection{Two Modes of Generation}
\label{sec:modes}

We distinguish evidence-backed generation (output supported by user-provided data, retrieved passages, explicit computation, or citations) from prior-only generation (output from the model's learned distribution without evidential grounding in the current context). Both modes are necessary; failure occurs when prior-only generation is emitted as if it were evidence-backed.

In this framing, the central issue is not generation itself. Both human cognition and language models are generative systems: they produce candidate content from prior structure before the environment fully determines the response. The relevant failure arises when internally generated content is granted more evidential authority than its available support warrants. In language models, this occurs when prior-driven completion is emitted as though it were sufficiently grounded in the current evidence rather than merely generated from learned continuation dynamics.

\subsection{The Misclassification Loop}
\label{sec:misclass}

A more precise characterization of hallucination is that the system produces output \emph{as if} it were sufficiently supported even when the available evidence is weak, absent, or unstable. The central problem is therefore not only incorrectness at the endpoint, but failure at the boundary where internally generated completion is allowed to become committed response.

This boundary problem can be stated as a simple functional sequence:
\begin{equation}
\text{internally generated content}
\;\rightarrow\;
\text{limited or ambiguous support}
\;\rightarrow\;
\text{misclassification of evidential sufficiency}
\;\rightarrow\;
\text{persistence of error-prone output}.
\label{eq:template}
\end{equation}

On this view, hallucination is a classification failure in which prior-driven completion crosses an evidential threshold it has not sufficiently earned. The model does not fail because it generates candidate continuations; it fails because the system surrounding generation allows weakly supported continuation to acquire the functional status of answer.

In biological predictive-processing models \citep{clark2013whatever, friston2010freeenergy}, a parallel failure mode arises when internally generated signals are classified as externally grounded evidence rather than as indicators of simulation-dominant processing. The LLM system lacks an analogous mode classifier at the output boundary. Importantly, this misclassification can be self-reinforcing: a model that produces output confidently---high self-consistency, stable paraphrase, apparent grounding in context---signals to downstream systems that it has earned outward commitment, even when that confidence reflects internal coherence rather than external warrant.

This framing yields three practical predictions. First, instruction-based refusal will be unreliable when the model's own self-assessment is part of the failure process. Second, external structural gating can catch cases where the model's self-classification fails. Third, the two mechanisms will exhibit complementary failure modes, motivating a composite abstention architecture rather than reliance on either alone.

\subsection{Why Abstention Is the Correct Intervention}
\label{sec:abstention_rationale}

The relevant intervention is classificatory rather than generative: the system must distinguish between content that is merely internally generated and content that is sufficiently supported for outward commitment. Abstention is therefore not a secondary safety heuristic in the present framework---it is the primary control point. Where support is insufficient, the correct action is interruption of unjustified commitment before emission, not post hoc repair of already emitted content.

This also clarifies why post hoc correction is inadequate as a primary strategy. Once unsupported content has been emitted, the system has already acted as though the content merited outward commitment. Correction after the fact may reduce harm from a specific output, but it does not address the underlying classification failure that allowed the boundary to be crossed.

The composite architecture evaluated in this paper is designed around this principle. Instruction-based refusal and structural gating are complementary ways of interrupting inappropriate outward commitment under uncertainty, each targeting a different point of failure in the evidential classification process.

\section{Method}
\label{sec:method}

\subsection{Black-Box Support-Deficit Score}
\label{sec:modescore}

The support-deficit score uses only signals computable without access to model internals.

\paragraph{Self-consistency ($A_t \in [0,1]$).} $K=3$ independent responses are generated; $A_t$ is the majority-vote agreement fraction. All generations used temperature $T=0.7$ and top-$p=0.9$.

\paragraph{Paraphrase stability ($P_t \in [0,1]$).} The query is rephrased and resubmitted; $P_t$ measures semantic overlap between original and rephrased responses.

\paragraph{Citation coverage ($C_t \in [0,1]$).} Fraction of content words in the response traceable to provided context. $C_t = 0$ when no context is provided. Operationally, $C_t$ is computed via keyword overlap between response content words and the provided context text (a proxy for attribution rather than entailment).

\paragraph{Support deficit:}
\begin{equation}
S_t = 1 - \frac{A_t + P_t + C_t}{3}
\end{equation}

\subsection{Abstention Policy}
\label{sec:policy}

\begin{equation}
\text{output}_t = 
\begin{cases} 
\text{answer} & \text{if } S_t \leq \tau, \\
\text{ABSTAIN} & \text{if } S_t > \tau,
\end{cases}
\end{equation}
where $\tau = 0.55$.

\subsection{Conditions}
\label{sec:conditions}

\begin{enumerate}
    \item \textbf{Baseline.} Standard generation, no abstention instructions.
    \item \textbf{Instruction-only.} System prompt instructs abstention when evidence is insufficient. No structural enforcement.
    \item \textbf{Hard-gated.} Baseline prompt; structural gate blocks output when $S_t > \tau$.
    \item \textbf{Composite.} Instruction prompt plus structural gate. Output is blocked if the model's own refusal fires or $S_t > \tau$. This tests whether combining the two mechanisms covers their individual failure modes.
\end{enumerate}

\subsection{Evaluation Design}
\label{sec:eval}

\subsubsection{Controlled Regime Evaluation (50 items)}
\label{sec:eval_regime}

Fifty items across five regimes (10 per regime) were evaluated under four conditions across three models, comprising 600 model-condition-item trials in total.

\begin{itemize}
    \item \textbf{R1: Answerable} ($n=10$). Factual queries with sufficient context.
    \item \textbf{R2: Unanswerable} ($n=10$). Context is empty or does not contain the answer.
    \item \textbf{R3: Conflicting evidence} ($n=10$). Two sources provide contradictory answers.
    \item \textbf{R4: Degraded retrieval} ($n=10$). Topically related context missing the answer.
    \item \textbf{R5: Adversarial pressure} ($n=10$). Unanswerable queries with instructions demanding confident answers.
\end{itemize}

Three models spanning a range of capability levels: GPT-3.5-turbo, GPT-4o-mini, and GPT-4o.

\subsubsection{TruthfulQA No-Context Abstention Stress Test (100 items)}
\label{sec:truthfulqa_method}

To evaluate abstention enforcement at larger scale under a uniform epistemic condition, we constructed a no-context stress test from TruthfulQA \citep{lin2022truthfulqa}. We sampled 100 items from the TruthfulQA ``generation'' validation split using a fixed random seed (seed=123) and presented each query with an empty context field. Under this configuration, no evidence is available, and the operationally correct behavior is to abstain; any non-abstaining response is counted as a hallucination. This stress test therefore measures abstention enforcement rather than factual accuracy on TruthfulQA.

The same three models (GPT-4o-mini, GPT-4o, GPT-3.5-turbo) were evaluated under the same four conditions. Gate parameters were $K=3$ and $\tau=0.55$. All generations used temperature $T=0.7$ and top-$p=0.9$.

\paragraph{Abstention detection.}
An output was classified as abstention if the normalized response was equal to or began with \texttt{abstain}, capturing punctuation and formatting variants (e.g., ``ABSTAIN.'').

\subsection{Metrics}
\label{sec:metrics}

\begin{itemize}
    \item \textbf{Accuracy:} correct answer OR correct abstention.
    \item \textbf{Hallucination rate:} answered when should have abstained.
    \item \textbf{Abstention rate:} declined to answer.
\end{itemize}

For the TruthfulQA stress test, because all 100 items have \texttt{should\_abstain=True}, correct\% equals abstain\% and hallucination\% equals the rate of answering despite absent evidence.

\section{Results}
\label{sec:results}

We report results in two parts. First, we analyze complementary failure modes in a controlled 50-item evaluation across five epistemic regimes (R1--R5). Second, we evaluate abstention enforcement under absent evidence using a 100-item no-context stress test derived from TruthfulQA.

\subsection{Overall Performance (50-item Regime Evaluation)}
\label{sec:overall}

Table~\ref{tab:overall} reports overall accuracy, hallucination, and abstention across models and conditions.

\begin{table}[t]
\caption{Overall performance across 50 items per model. Bold indicates best performance per model.}
\label{tab:overall}
\begin{center}
\begin{tabular}{llccc}
\toprule
Model & Condition & Accuracy & Hallucination & Abstention \\
\midrule
\multirow{4}{*}{GPT-4o-mini} 
& 1. Baseline & 62\% & 38\% & 42\% \\
& 2. Instruction & 98\% & 0\% & 82\% \\
& 3. Gate & 80\% & 20\% & 60\% \\
& 4. Composite & \textbf{98\%} & \textbf{0\%} & 82\% \\
\midrule
\multirow{4}{*}{GPT-4o} 
& 1. Baseline & 70\% & 30\% & 50\% \\
& 2. Instruction & 98\% & 0\% & 82\% \\
& 3. Gate & 82\% & 18\% & 62\% \\
& 4. Composite & \textbf{98\%} & \textbf{0\%} & 82\% \\
\midrule
\multirow{4}{*}{GPT-3.5-turbo} 
& 1. Baseline & 50\% & 50\% & 30\% \\
& 2. Instruction & 94\% & 6\% & 76\% \\
& 3. Gate & 76\% & 24\% & 56\% \\
& 4. Composite & \textbf{96\%} & \textbf{4\%} & 78\% \\
\bottomrule
\end{tabular}
\end{center}
\end{table}

The composite condition achieved 96--98\% accuracy with 0--4\% hallucination across all three models. Instruction-only already achieved 0\% hallucination on GPT-4o-mini and GPT-4o; the composite reduced GPT-3.5-turbo hallucination from 6\% to 4\% while preserving the same abstention pattern. The composite inherits the instruction component's behavior: for GPT-4o-mini and GPT-4o, this includes 10\% over-cautious abstention on answerable items; for GPT-3.5-turbo, it retains residual vulnerability on conflicting-evidence items.

\subsection{TruthfulQA-derived No-Context Abstention Stress Test (100 items)}
\label{sec:truthfulqa}

Table~\ref{tab:truthfulqa} reports results on the 100-item TruthfulQA no-context stress test. In this setting, all items require abstention; ``Correct'' denotes correct abstention and ``Halluc'' denotes answering despite absent evidence.

\begin{table}[t]
\caption{TruthfulQA 100-item no-context abstention stress test. ``Correct'' denotes correct abstention (output begins with \texttt{ABSTAIN}); ``Halluc'' denotes answering despite absent evidence. These results quantify abstention enforcement, not TruthfulQA benchmark performance.}
\label{tab:truthfulqa}
\centering
\begin{tabular}{llccc}
\toprule
Model & Condition & Correct & Halluc & Abstain \\
\midrule
\multirow{4}{*}{GPT-4o-mini}
& Baseline & 0\% & 100\% & 0\% \\
& Instruction & 100\% & 0\% & 100\% \\
& Gate & 98\% & 2\% & 98\% \\
& Composite & 98\% & 2\% & 98\% \\
\midrule
\multirow{4}{*}{GPT-4o}
& Baseline & 0\% & 100\% & 0\% \\
& Instruction & 100\% & 0\% & 100\% \\
& Gate & 99\% & 1\% & 99\% \\
& Composite & 100\% & 0\% & 100\% \\
\midrule
\multirow{4}{*}{GPT-3.5-turbo}
& Baseline & 0\% & 100\% & 0\% \\
& Instruction & 62\% & 38\% & 62\% \\
& Gate & 100\% & 0\% & 100\% \\
& Composite & 100\% & 0\% & 100\% \\
\bottomrule
\end{tabular}
\end{table}

Baseline generation produced 0\% abstention across all three models. Instruction-only abstention was capability-sensitive: GPT-4o and GPT-4o-mini abstained on all 100 items, whereas GPT-3.5-turbo abstained on 62 of 100 items and answered 38. The hard gate enforced 98--100\% abstention across models regardless of instruction-following capability. The composite achieved 98--100\% abstention across models, with 2\% leakage on GPT-4o-mini.

The GPT-3.5-turbo results illustrate the capability dependence of instruction-based abstention: the same verbal instruction that produced 100\% abstention for GPT-4o and GPT-4o-mini produced only 62\% for GPT-3.5-turbo. The structural gate compensated for this gap, enforcing 100\% abstention for GPT-3.5-turbo in both the gate and composite conditions.

\subsection{Performance by Regime}
\label{sec:byregime}

Table~\ref{tab:regime} reports composite-condition performance by regime and model.

\begin{table}[t]
\caption{Composite condition accuracy and hallucination by regime across models.}
\label{tab:regime}
\begin{center}
\begin{tabular}{lcccccc}
\toprule
& \multicolumn{2}{c}{GPT-4o-mini} & \multicolumn{2}{c}{GPT-4o} & \multicolumn{2}{c}{GPT-3.5-turbo} \\
\cmidrule(lr){2-3} \cmidrule(lr){4-5} \cmidrule(lr){6-7}
Regime & Acc & Hall & Acc & Hall & Acc & Hall \\
\midrule
R1: Answerable & 90\% & 0\% & 90\% & 0\% & 90\% & 10\% \\
R2: Unanswerable & 100\% & 0\% & 100\% & 0\% & 100\% & 0\% \\
R3: Conflicting & 100\% & 0\% & 100\% & 0\% & 90\% & 10\% \\
R4: Degraded & 100\% & 0\% & 100\% & 0\% & 100\% & 0\% \\
R5: Adversarial & 100\% & 0\% & 100\% & 0\% & 100\% & 0\% \\
\bottomrule
\end{tabular}
\end{center}
\end{table}

The composite achieved 100\% accuracy on Regimes 2--5 for GPT-4o-mini and GPT-4o. GPT-3.5-turbo showed residual hallucination on R1 (10\%, one item answered incorrectly) and R3 (10\%, one undetected conflict), consistent with its weaker instruction-following capability.

\subsection{Why Neither Mechanism Alone Suffices}
\label{sec:complementary}

The central finding of the present experiments is that near-zero hallucination required combining complementary mechanisms. Table~\ref{tab:complementary} illustrates this for GPT-4o-mini.

\begin{table}[t]
\caption{Complementary failure modes (GPT-4o-mini). Each mechanism fails where the other succeeds.}
\label{tab:complementary}
\begin{center}
\begin{tabular}{lcccc}
\toprule
Regime & Baseline & Instruction & Gate & Composite \\
\midrule
R1: Answerable (accuracy) & 100\% & 90\% & 100\% & 90\% \\
R3: Conflicting (halluc.) & 100\% & 0\% & 70\% & 0\% \\
R4: Degraded (halluc.) & 30\% & 0\% & 20\% & 0\% \\
R5: Adversarial (halluc.) & 50\% & 0\% & 10\% & 0\% \\
\bottomrule
\end{tabular}
\end{center}
\end{table}

\paragraph{Gate failure mode: confident confabulation.} On Regime 3 (conflicting evidence), the gate hallucinated on 70\% of items across GPT-4o-mini and GPT-4o. Examination of support-deficit scores reveals the mechanism: the model selected one source's answer and produced it with high self-consistency ($A_t \geq 0.67$), high paraphrase stability ($P_t \geq 0.46$), and high citation coverage ($C_t \geq 0.26$, since the chosen answer appeared in context). The support deficit $S_t$ remained below threshold because the model was confidently wrong---consistent, stable, and grounded in one side of the conflict.

\paragraph{Instruction failure mode: over-cautious abstention and residual hallucination.} Instruction-only produced 10\% incorrect abstentions on R1 for GPT-4o-mini and GPT-4o; GPT-3.5 instead produced 10\% hallucination on one answerable item. The gate produced 0\% incorrect abstentions on R1 across all models, because the support-deficit score correctly identified high $A_t$, $P_t$, and $C_t$ on answerable items.

\paragraph{Composite resolution.} The composite combines both mechanisms via logical OR: output is blocked if the instruction-based model refuses or $S_t > \tau$. On R3, the instruction component can catch some source-conflict cases that the gate misses, even when the final generation appears internally coherent. On R1, the gate answers all items correctly, but the instruction component abstains on one R1 item for GPT-4o-mini and GPT-4o; the composite inherits that abstention. For GPT-3.5-turbo, the composite improves upon instruction-only by catching some conflicting-evidence cases the instruction component missed.

\subsection{Cross-Model Consistency}
\label{sec:consistency}

The pattern is stable across models. All three models show: (a) baseline hallucination of 30--50\%; (b) instruction-only reducing overall hallucination to 0\% for GPT-4o-mini and GPT-4o and to 6\% for GPT-3.5-turbo, with 10\% over-cautious abstention on answerable items for GPT-4o-mini and GPT-4o; (c) the gate alone reducing hallucination to 18--24\% with 0\% incorrect abstentions on answerable items; and (d) the composite achieving 0--4\% hallucination overall. The consistency across these three models suggests the pattern is stable within the tested OpenAI model family, though architecture-level generality remains to be established.

GPT-3.5-turbo showed slightly worse performance: instruction-only hallucinated on 6\% of items (vs.\ 0\% for the larger models), and the composite retained 4\% residual hallucination. These failures were concentrated in the conflicting-evidence regime, where weaker instruction-following allowed the model to bypass the verbal abstention instruction more often than GPT-4o-mini/4o.

The TruthfulQA stress test reinforces this pattern. Under a uniform no-context condition, instruction-only abstention degraded from 100\% (GPT-4o, GPT-4o-mini) to 62\% (GPT-3.5-turbo), while the structural gate maintained 98--100\% abstention across all models. This confirms that instruction-based mechanisms degrade with model capability, while the structural gate provides a capability-independent abstention floor.

\subsection{Hypothesis Evaluation}
\label{sec:hypotheses}

\begin{itemize}
    \item \textbf{H1 (Hallucination reduction):} Supported. The composite reduced hallucination from 30--50\% (baseline) to 0--4\% across all models.
    \item \textbf{H2 (Accuracy preservation):} Partially supported. The gate alone achieved 100\% R1 accuracy across all models (0\% incorrect abstentions). The composite achieved 90\% for GPT-4o-mini/4o (due to instruction component's over-caution) and 90\% for GPT-3.5 (due to residual hallucination).
    \item \textbf{H3 (Gate outperforms instruction on $\geq$1 regime):} Supported. On R1, the gate (100\%) outperformed instruction (90\%) on all three models.
    \item \textbf{H4 (Adversarial robustness):} Supported. The composite achieved 100\% correct abstention on R5 across all models.
    \item \textbf{H5 (Capability-independent abstention floor):} Supported by the TruthfulQA stress test. The structural gate enforced 98--100\% abstention across all three models, whereas instruction-only abstention was 62\% for GPT-3.5-turbo.
\end{itemize}

\section{Discussion}
\label{sec:discussion}

\subsection{The Case for Composite Architectures}
\label{sec:composite}

The central finding of the present experiments is that near-zero hallucination required combining complementary mechanisms. Instruction-based refusal leverages the model's internal representations of evidential status---effective when those representations are accurate, but unreliable when the model is confidently wrong (R3, R4) or when instruction-following degrades (GPT-3.5-turbo). Structural gating bypasses the model's self-assessment using external signals---effective at detecting uncertainty and adversarial pressure, but blind to confident confabulation where the model produces consistent, stable, grounded-looking output from parametric memory.

The composite combines the advantages of both mechanisms while also inheriting some of the instruction component's abstention behavior on answerable items, especially for GPT-4o-mini and GPT-4o. The instruction component can catch source-conflict cases the gate misses, while structural gating catches cases where instruction-following fails and provides a capability-independent safety floor.

\subsection{The Confident-Confabulation Boundary}
\label{sec:boundary}

The gate's failure on Regime 3 identifies a principled boundary of black-box signal detection. Self-consistency measures internal agreement, not correspondence to external evidence. A model that consistently selects one side of a conflict produces high $A_t$ and high $P_t$---indistinguishable, on these signals, from a model that correctly answers a well-grounded question. The verdict was stable; the grounding was not.

This suggests that the support-deficit signal set should be extended with an explicit source-conflict detection signal: checking whether the context contains contradictory evidence, independent of the model's answer. This could be implemented as a lightweight entailment check or a structured prompt that asks the model to identify disagreements in the source material before answering.

\subsection{Endpoint Correctness Is Not Support Stability}
\label{sec:endpoints}

A broader implication of the output-boundary framing is that endpoint correctness may not adequately capture support stability. A model can preserve a nominally acceptable answer while relying on weakly grounded or procedurally unstable intermediate reasoning. This suggests that hallucination evaluation should not be restricted to output correctness alone, but should also consider whether the system crossed the evidential boundary appropriately when deciding to emit.

This concern is visible in the R3 results: models achieved internal consistency and stable paraphrase while selecting one side of a genuine conflict. Evaluation that rewards only final-answer accuracy would overestimate reliability in exactly this class of case.

\subsection{Relation to Predictive Processing}
\label{sec:biology}

The complementary failure modes map onto principles from predictive processing and control-theoretic models of biological inference \citep{clark2013whatever, friston2010freeenergy}. In these frameworks, the misclassification loop is hardest to break when the internally generated signal closely resembles genuine external evidence. The gate fails for exactly this reason: confident confabulation mimics evidence-backed generation on all observable signals. This is consistent with the finding that the most effective intervention combines a support-deficit classifier (structural gate) with instruction-based source assessment---precisely the composite architecture tested here.

\subsection{Practical Implications}
\label{sec:practical}

The composite architecture requires approximately $6K + 4$ API calls per query ($K$ consistency samples, 2 paraphrase probes, 1 gated generation, 1 instructed generation). At $K=3$, this is 22 calls per query---a non-trivial cost that must be weighed against the application's tolerance for hallucination. For high-stakes domains (medical, legal, financial), the cost is justified. For casual conversation, it is not.

The 10\% over-cautious abstention rate on answerable items for GPT-4o-mini and GPT-4o (inherited from the instruction component) represents a coverage-accuracy tradeoff. In deployment, this could be addressed by asymmetric thresholds: requiring both mechanisms to agree on abstention for answerable-type queries, or using the gate to override instruction-based refusal when $S_t$ is very low.

\subsection{Limitations}
\label{sec:limitations}

\paragraph{Scale.} The controlled evaluation used 50 items per model, 10 per regime. The TruthfulQA stress test used 100 items per model under a single epistemic condition (no context). Larger-scale evaluation on established benchmarks (SQuAD 2.0, HaluEval) with greater sample sizes is needed for publication-grade statistical power.

\paragraph{TruthfulQA scope.} TruthfulQA was used exclusively in an evidence-free (empty-context) stress test to evaluate abstention enforcement. Results quantify the rate at which each condition correctly withholds output when no evidence is available; they do not measure answer correctness on TruthfulQA and should not be interpreted as TruthfulQA benchmark performance.

\paragraph{Model family.} All three models are from OpenAI. Cross-family evaluation (Llama, Claude, Mistral) would strengthen generalization claims. Claims about autoregressive generation in this paper are limited to this model family and may not generalize to other architectures (e.g., different RLHF stacks, safety tuning, or decoding defaults).

\paragraph{Simplified signals.} Citation coverage uses keyword overlap rather than entailment. Paraphrase stability uses surface-level word overlap. $C_t$ is not a factuality verifier; it is one signal among three in a composite deficit score. Stronger signal implementations would likely improve performance.

\paragraph{Equal weighting.} The three signals are equally weighted in $S_t$. Learned weights calibrated on a development set could improve threshold discrimination.

\paragraph{Scope.} Restricted to factual QA. Open-ended generation, reasoning chains, and multi-turn dialogue are not addressed.

\paragraph{Synthetic regime construction.} All five epistemic regimes (R1--R5) were hand-constructed for controlled evaluation rather than sampled from natural query distributions. This enables precise failure-mode analysis but risks implicit prompt leakage or regime overfitting. These regimes are designed to isolate specific epistemic conditions, not to approximate real-world query distributions.

\paragraph{Dependence on sampling stochasticity.} Self-consistency ($A_t$) relies on stochastic sampling; under deterministic decoding (temperature=0), $A_t$ may collapse to 1.0 for all items, eliminating the signal. We used $T=0.7$ and top-$p=0.9$ for all experiments but did not vary decoding strategy systematically.

\paragraph{Cost and deployment feasibility.} The composite architecture requires $\sim$22 API calls per query ($K=3$). This is presented as a research architecture demonstrating mechanism complementarity, not as a production-ready solution.

\paragraph{Instruction prompt sensitivity.} Instruction-only performance depends on prompt wording, which we did not vary. No robustness analysis over paraphrased instructions or adversarial prompt engineering was conducted.

\paragraph{Scope of the theoretical bridge.} The present paper does not claim that all hallucination reduces to a single predictive principle, nor that language models and human cognition share identical mechanisms. The proposed overlap between the output-boundary framing and predictive-processing accounts of misclassification is functional rather than ontological. The empirical study is offered as a bounded proof of concept for a support-classification framework, not as final confirmation of a universal account of hallucination.

\subsection{Disconfirmation Criteria}
\label{sec:disconfirm}

\begin{itemize}
    \item \textbf{Gate value:} If larger-scale replication shows the composite does not reduce hallucination below instruction-only, the gate component adds no value.
    \item \textbf{R1 advantage:} If instruction-only achieves 0\% incorrect abstentions on answerable items across models, the gate's R1 advantage disappears.
    \item \textbf{Family generality:} If the pattern does not hold across model families (Llama, Claude), the mechanism is OpenAI-specific.
    \item \textbf{Temperature:} If temperature=0 baseline achieves comparable results, the multi-sample support-deficit score is unnecessary overhead.
    \item \textbf{Prompt robustness:} If the pattern does not hold when varying instruction prompts, the composite's reliance on verbal instruction is fragile.
    \item \textbf{Decoding:} If temperature=0 eliminates gate effectiveness (no variance for self-consistency), the multi-sample architecture is required.
\end{itemize}

\section{Conclusion}
\label{sec:conclusion}

We proposed that LLM hallucination is a misclassification error at the output boundary and evaluated four mitigation strategies across three models and five epistemic regimes. The central finding of the present experiments is that near-zero hallucination required combining complementary mechanisms: neither instruction-based refusal nor structural gating alone was sufficient. The composite architecture achieved 96--98\% accuracy with 0--4\% hallucination across models spanning a substantial capability range. A supplementary 100-item no-context stress test derived from TruthfulQA confirmed that the structural gate provides a capability-independent abstention floor: instruction-only abstention degraded to 62\% for GPT-3.5-turbo, while the gate and composite maintained 98--100\% abstention across all models.

The gate's specific failure---confident confabulation, where the model is consistently and stably wrong---identifies a principled boundary of black-box signal detection and motivates extension with source-conflict detection signals. The instruction component's specific failure---over-cautious abstention on answerable items for GPT-4o-mini/4o and residual hallucination for GPT-3.5---identifies the limits of verbal self-regulation and motivates structural override. The observation that endpoint correctness can coexist with support instability (most clearly in R3) suggests that hallucination evaluation should attend to boundary control as well as final-answer accuracy.

These complementary failure modes are consistent with the control-theoretic framing: a system that generates internal predictions requires both a support-deficit classifier (structural gate) and instruction-based source assessment. Neither alone is a complete solution; together they approach one. We present these results as initial evidence for a composite approach within the OpenAI model family and invite replication at larger scale, across model families, and on established benchmarks.

\section*{Data and Code Availability}
The evaluation prompts, controlled regime items (R1--R5), and supporting code used in this study are included in the supplementary materials accompanying this manuscript. A public reproducibility package including prompts, scoring scripts, and run logs will be released upon publication.

\printbibliography

\appendix

\section{Reproducibility Checklist for JAIR}

\subsection*{All articles:}
\begin{enumerate}
    \item All claims investigated in this work are clearly stated. [yes]
    \item Clear explanations are given how the work reported substantiates the claims. [yes]
    \item Limitations or technical assumptions are stated clearly and explicitly. [yes]
    \item Conceptual outlines and/or pseudo-code descriptions of the AI methods introduced in this work are provided, and important implementation details are discussed. [yes]
    \item Motivation is provided for all design choices, including algorithms, implementation choices, parameters, data sets and experimental protocols beyond metrics. [yes]
\end{enumerate}

\subsection*{Articles containing theoretical contributions:}
Does this paper make theoretical contributions? [yes]

\begin{enumerate}
    \item All assumptions and restrictions are stated clearly and formally. [partially]
    \item All novel claims are stated formally (e.g., in theorem statements). [no]
    \item Proofs of all non-trivial claims are provided in sufficient detail to permit verification. [NA]
    \item Complex formalism, such as definitions or proofs, is motivated and explained clearly. [yes]
    \item The use of mathematical notation and formalism serves the purpose of enhancing clarity and precision. [yes]
    \item Appropriate citations are given for all non-trivial theoretical tools and techniques. [yes]
\end{enumerate}

\subsection*{Articles reporting on computational experiments:}
Does this paper include computational experiments? [yes]

\begin{enumerate}
    \item All source code required for conducting experiments is included in an online appendix or will be made publicly available upon publication. [yes]
    \item The source code comes with a license that allows free usage for reproducibility purposes. [partially]
    \item The source code comes with a license that allows free usage for research purposes in general. [partially]
    \item Raw, unaggregated data from all experiments is included in an online appendix or will be made publicly available upon publication. [yes]
    \item The unaggregated data comes with a license that allows free usage for reproducibility purposes. [partially]
    \item The unaggregated data comes with a license that allows free usage for research purposes in general. [partially]
    \item If an algorithm depends on randomness, then the method used for generating random numbers and for setting seeds is described sufficiently to allow replication. [yes]
    \item The execution environment for experiments, including hardware, software, and library versions, is described. [partially]
    \item The evaluation metrics used in experiments are clearly explained and their choice is explicitly motivated. [yes]
    \item The number of algorithm runs used to compute each result is reported. [yes]
    \item Reported results have not been cherry-picked by silently ignoring unsuccessful or unsatisfactory experiments. [yes]
    \item Analysis of results goes beyond single-dimensional summaries of performance. [yes]
    \item All (hyper-)parameter settings for the algorithms/methods used in experiments have been reported, along with the rationale or method for determining them. [yes]
    \item The number and range of (hyper-)parameter settings explored prior to conducting final experiments have been indicated. [partially]
    \item Appropriately chosen statistical hypothesis tests are used to establish statistical significance in the presence of noise effects. [no]
\end{enumerate}

\subsection*{Articles using data sets:}
Does this work rely on one or more data sets? [yes]

\begin{enumerate}
    \item All newly introduced data sets are included in an online appendix or will be made publicly available upon publication. [yes]
    \item The newly introduced data set comes with a license that allows free usage for reproducibility purposes. [partially]
    \item The newly introduced data set comes with a license that allows free usage for research purposes in general. [partially]
    \item All data sets drawn from the literature or other public sources are accompanied by appropriate citations. [yes]
    \item All data sets drawn from the existing literature are publicly available. [yes]
    \item All new data sets and data sets that are not publicly available are described in detail. [yes]
    \item All methods used for preprocessing, augmenting, batching or splitting data sets are described in detail. [yes]
\end{enumerate}

\subsection*{Explanations on any of the answers above (optional):}
The paper presents a controlled proof-of-concept evaluation of a composite abstention architecture across hand-constructed epistemic regimes and a TruthfulQA-derived no-context stress test. Source code, prompts, evaluation items, and run logs are provided in the supplementary materials. Statistical significance testing was not performed because the goal of the experiments was mechanism demonstration rather than benchmark-scale comparison.

\end{document}